\pdfoutput=1
\documentclass[11pt]{article}
\usepackage[final]{acl}

\usepackage{times}
\usepackage{latexsym}

\usepackage[T1]{fontenc}
\usepackage[utf8]{inputenc}
\usepackage{microtype}
\usepackage{inconsolata}
\usepackage{graphicx}
\usepackage{amsmath}

\usepackage{booktabs}
\usepackage{multirow}
\usepackage{tabularx}
\usepackage{colortbl}
\usepackage{arydshln}
\usepackage{subcaption}
\usepackage{graphicx}

\usepackage{todonotes}

\usepackage{amssymb}
\usepackage{cleveref}
\usepackage{bm}
\crefname{section}{§}{§§}
\Crefname{section}{§}{§§}

\title{Entropy-Based Decoding for Retrieval-Augmented Large Language Models}

\author{Zexuan Qiu$^1$\thanks{Correspondence to: zxqiu22@cse.cuhk.edu.hk. } \quad Zijing Ou$^2$\quad Bin Wu$^3$\quad Jingjing Li$^1$\quad Aiwei Liu$^4$\quad Irwin King$^1$ \vspace{1.5mm}\\
$^1$The Chinese University of Hong Kong\quad $^2$Imperial College London \\
$^3$University College London\quad $^4$Tsinghua University \\
}

\newcommand{\modela}{LeEns}
\newcommand{\modelb}{CLeHe}
\newcommand{\replug}{\textsc{RePlug}}
\newcommand{\naive}{\textsc{Naive}}
\newcommand{\avgens}{AvgEns}
\newcommand{\llama}{\textsc{Llama}}

\newcommand{\ie}{\textit{i.e.}}
\newcommand{\eg}{\textit{e.g.}}

\begin{document}
\maketitle
\begin{abstract}
Augmenting Large Language Models (LLMs) with retrieved external knowledge has proven effective in improving the factual accuracy of generated responses. Despite their success, retrieval-augmented LLMs still face the distractibility issue, where the generated responses are negatively influenced by noise from both external and internal knowledge sources. In this paper, we introduce a novel, training-free decoding method guided by entropy considerations to mitigate this issue. Our approach utilizes entropy-based document-parallel ensemble decoding to prioritize low-entropy distributions from retrieved documents, thereby enhancing the extraction of relevant information of context. Additionally, it incorporates a contrastive decoding mechanism that contrasts the obtained low-entropy ensemble distribution with the high-entropy distribution derived from the model's internal knowledge across layers, which ensures a greater emphasis on reliable external information. Extensive experiments on open-domain question answering datasets demonstrate the superiority of our method.\footnote{Our code is available at \url{https://github.com/zexuanqiu/entropy-based-decoding}.}
\end{abstract}
\section{Introduction}
\vspace{-0.5mm}
In recent years, Large language models (LLMs) have revolutionized natural language processing, showcasing remarkable performance across various downstream tasks~\citep{brown2020language,ouyang2022training,touvron2023llama}. However, they still struggle with hallucination due to the inaccuracy of parametric memory~\citep{bubeck2023sparks} and inherently tend to produce outdated information~\citep{kasai2024realtime}. In contrast, explicitly augmenting LLMs with retrieved external knowledge from reliable datastores~\citep{lewis2020retrieval,borgeaud2022improving} can enable LLMs to generate content that exhibits less deviation from the truth, and benefit downstream knowledge-intensive tasks~\citep{petroni2020kilt}.

Despite the success of retrieval-augmented LLMs, the augmented generation is still sub-optimal due to the \textit{distractibility issue}, where the generated responses are easily negatively affected by noise from both external knowledge and intrinsic model knowledge.  As for the input context, LLMs' understanding of context can be \textit{explicitly distracted by irrelevant parts} within the retrieved context~\citep{shi2023large}. A typical illustrative case is the ``lost in the middle'' distraction phenomenon observed in the synthetic multi-document question-answering scenario~\citep{liu2024lost, qiu2024clongeval}, where the oracle document containing the correct answer is encircled by numerous retrieved distracting documents. In this scenario, LLMs frequently fail to deliver the correct answer unless the oracle document is strategically placed at the very beginning or end of the context. With regard to intrinsic knowledge, LLMs are easily \textit{implicitly distracted by the parametric knowledge} acquired during pre-training. This is in conflict with retrieval-augmented generation which is expected to generate responses based on reliable retrieved context.  Particularly in the domain of question answering, previous works~\citep{longpre2021entity,xie2023adaptive} show that LLMs stubbornly adhere to their built-in knowledge even when it conflicts with external knowledge. 

How to eliminate the impact of the above-mentioned distractibility issue, so as to extract useful knowledge from the retrieved context for the input query, is our research focus. Although existing works strive to effectively leverage the retrieved context by directly fine-tuning retrieval-augmented LLMs~\citep{lin2023ra, zhang2024raft,li2024retrieval} or incorporating trainable encoder modules~\citep{izacard2020leveraging, yen2024long}, these approaches require additional training, rendering them potentially impractical in resource-constrained environments. In this paper, we propose a novel decoding method guided by entropy considerations to simultaneously mitigate the impact of noisy information from both the external context and parametric knowledge. 
The proposed method can be seamlessly integrated into LLMs without requiring additional tuning.

Specifically, to enhance LLMs’ ability to extract useful information from multiple retrieved documents, we let LLMs process each retrieved document in parallel and ensemble of the output distributions from each document to determine the next-token distribution, with the ensemble weights adaptively assigned based on the uncertainty of each document-conditioned distribution. At each generation step, documents with lower uncertainty (\ie, lower entropy) in the LLM output are given more attention during decoding. Ultimately, we obtain a low-entropy distribution aggregated among documents. Furthermore, to alleviate the potential distraction from parametric knowledge, we refine the next-token distribution by contrasting the obtained low-entropy distribution when feeding the retrieved documents, against the distribution without context. Here, we propose to use the distribution from the layer exhibiting the highest entropy without context for contrast, in order to highlight the proportional changes in token probabilities after introducing external knowledge. 

The proposed decoding method shows an impressive performance in the synthetic challenging multi-document scenario~\citep{liu2024lost} where the negative impact of retrieved distractor documents is emphasized. We further conduct extensive experiments across four LLMs of varying sizes on four diverse open-domain question answering tasks including NQ~\citep{kwiatkowski2019natural}, TriviaQA~\citep{JoshiCWZ17}, WebQ~\citep{berant2013semantic} and PopQA~\citep{MallenAZDKH23}. Experimental results confirm the superiority of our methods and validate the effectiveness of each component.

\begin{figure*}[ht]
    \centering
    \includegraphics[width=\linewidth]{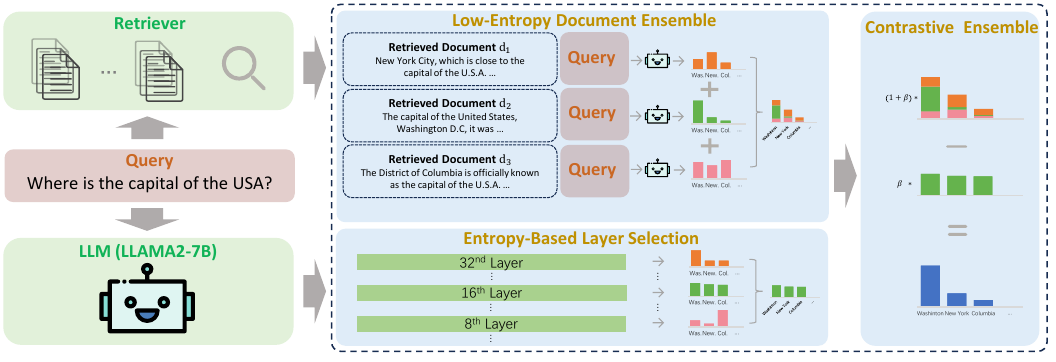}
    \caption{Overview of the decoding process of \modelb.}
    \label{fig:method-illustration}
    \vspace{-4.5mm}
\end{figure*}

\section{Methodology}
In this work, we will investigate a new decoding approach for retrieval-augmented generation (RAG). Given an input query $x$, RAG first retrieves top-$K$ relevant documents $D := \{d_1, d_2, \dots, d_K\}$ from the knowledge base via a retriever as external evidence, which is then incorporated with the query as the input to large language model parametrized by $\theta$ for generating a faithful response $y$. 

Despite the integration of external knowledge, the decoding manner does matter to achieve desirable performance in RAG. A common approach, termed as ``\naive~RAG'', involves concatenating the query $x$, the previously generated response $y_{<t}$, and the retrieved documents $D$ as the input sequence, resulting  in the following decoding method
\begin{equation}
    p_\theta (y_t|D, x, y_{<t}) = p(y_t|d_0 \circ \cdots \circ d_k \circ x \circ y_{<t}),
    \label{fml:naive}
\end{equation}
where $\circ$ denotes the concatenation operation.
Even though this method has demonstrated superior performance, \citet{liu2024lost} highlight that it suffers from the ``lost in the middle'' distraction phenomenon, where LLMs tend to overlook the oracle document due to the distraction from other documents in $D$, unless the oracle document is placed at the very beginning or the end.
Moreover, this simple approach does not consider the potential negative effects of the underlying parametric knowledge of LLMs. In the subsequent sections, we discuss how to mitigate these two issues simultaneously by entropy considerations.

\subsection{Entropy-Based Document Ensemble}
\label{subsec: leens}
Instead of naively concatenating the documents $D$, we propose to alleviate the ``loss in the middle'' issue using the product-of-experts\footnote{Empirically, we find that using product-of-experts and mixture-of-experts methods yield similar performance.} ensemble approach~\cite{hinton2002training}. Specifically, we model the log probability of the next-token distribution as
\begin{equation}
\begin{aligned}
    \!\log p_\theta(y_t|D) 
    & \!\propto\!\! \sum_{j=1}^{K} \! w_{j,t}  \log p_\theta(y_t|d_j \!\circ\! x \!\circ\! y_{<t}), \!\!\!\!
\end{aligned}
\label{fml: poe}
\end{equation}
where $\sum_j w_{j,t} = 1, \forall t$, $p_\theta(y_t|D,x,y_{<t})$ is denoted as $p_\theta(y_t|D)$ for short and $w_{j,t}$ denotes the weight of the $j$-th document on generating the token at the $t$-th time step. In Eq.~\eqref{fml: poe}, each document in $D$ is concatenated with the query and the previously generated response. This combined input is then individually fed into the LLM. The output logit scores are subsequently averaged using the weights $w_{j,t}$. This ensemble approach, which leverages parallel decoding, helps mitigate position bias and provides a more effective means of utilizing the retrieved documents.

There are multiple choices to compute the weights. A straightforward option is to use uniform weighting, \ie, $w_{j,t} = \frac{1}{|D|}$. However, this method may fail to effectively extract valuable information when irrelevant documents are included among the top-$K$ retrieved documents.
Another option is to utilize a similarity score $s(d_j, x)$ between the query and the retrieved document, \eg, BM25, as the time-independent ensemble weights $w_{j,t} \propto s(d_j, x)$ \citep{lewis2020retrieval}. However, this retriever-based scoring approach may hinder the LLM's ability to extract relevant information when the retrieved documents contain numerous distractors, drastically reducing the factual accuracy of the responses as illustrated in \cref{subsec:middle}.

We posit that the uncertainty present in the next-token distribution inherently serves as a reliable indicator of the informativeness of the retrieved documents. Similar concepts have been employed in previous work to reduce hallucinations in LLMs~\citep{van2022mutual,varshney2023stitch}. Consequently, we propose using an entropy-based score $w_{j,t}^{ \textsc{h}}$ as the preference weight for each document at each decoding step:
\begin{equation}
\begin{aligned}
    w_{j,t}^{ \textsc{h}} & = \frac{\exp ^{(- H_{j,t}) /\tau}}{\sum_{d_k\in D} \exp ^{(- H_{j,t}) /\tau}}, \\ 
    H_{j,t} & = -\sum_{y_t \in \mathcal{V}}p_\theta(y_t|d_j) \log p_\theta(y_t|d_j),
\end{aligned}
\label{fml:entropy}
\end{equation}
where $p_\theta(y_t|d_j)$ denotes $p_\theta(y_t|d_j \! \circ \! x \!\circ \! y_{<t})$ for short, $\mathcal{V}$ represents the vocabulary set and $\tau$ is a hyper-parameter controlling the concentration level of distributions. The motivation behind Eq.~\eqref{fml:entropy} is that the LLM can autonomously evaluate the significance of each document during the generation process. Intuitively, it implies that those document-conditioned distributions with lower uncertainty will be assigned higher weights. Such a time-dependent approach can effectively capture useful information from the retrieved documents at each generation step, thereby influencing the generation process more significantly. We  refer to this method as \textbf{\modela}~(\textbf{L}ow-\textbf{e}ntropy \textbf{Ens}emble).

\subsection{Entropy-Based Contrastive Decoding}
\label{subsec: clehe}
While \textbf{\modela} can effectively help LLMs discern valuable evidence from external knowledge, the parametric knowledge of LLMs embedded during the pre-training phase might affect the answer generation, especially when these two types of knowledge conflict~\cite{longpre2021entity,xie2023adaptive}. In this section, we propose to address this issue via entropy-based contrastive decoding.

\paragraph{Contrastive Decoding with PMI.}
Inspired by the success in contrastive decoding to mitigate hallucination of LLMs \citep{shi2023trusting,li2022contrastive}, we adjust the logit score $z_t$ for the generated token $y_t \in \mathcal{V}$ at the $t$-th time step by incorporating the pointwise mutual information (PMI) between $y_t$ and the document set $D$, given the query $x$:
\vspace{-2mm}
\begin{equation}
\begin{aligned}
    \!\!\!\!\!\!\!z_t & = \log p_\theta^\textsc{h}(y_t|D)+ \beta \overbrace{\log \frac{p_\theta^\textsc{h}(y_t|D)}{p_\theta(y_t|x,y_{<t})}}^{\text{PMI}} \\
    & =\!\! (1 \!+\! \beta) \log p_\theta^\textsc{h}(y_t|D) \!-\! \beta \log p_\theta(y_t|x,y_{<t}),
\end{aligned}
\label{fml: add_pmi}
\end{equation}
where $\beta$ is a positive coefficient to control the contrast intensity, and $p_\theta^\textsc{h}(y_t|D)$ denotes the previously proposed entropy-based document ensemble distribution. Intuitively, PMI serves as a measurement of information gains. It is evident that the model tends to generate tokens with a high probability of $p_\theta^\textsc{h}(y_t|D)$ and a low probability of $p_\theta(y_t|x,y_{<t})$. These tokens provide greater information gain for the next token generation. Consequently, incorporating PMI can enhance the model's reliance on external knowledge.

\paragraph{Layer-wise Contrast with High Entropy.}
To perform contrastive decoding, it is necessary to compute 
$p_\theta(y_t|x, y_{<t})$. This can be achieved by taking the hidden states from the last layer of LLMs and passing them through the classification head.
However, the distribution derived from the last layer may exhibit overconfidence, characterized by extremely low probabilities for most words and disproportionately high probabilities for a few. 
Such overconfidence can erroneously amplify external knowledge when conducting contrasting, potentially leading to false positive failures, as illustrated in Figure~\ref{fig:app-false-positive} in the Appendix.

To address this issue, we propose selecting the layer that contains the most ``ambiguous'' parametric knowledge among the layers as a proper reference for contrast. This allows the model to more effectively leverage external knowledge, reducing overconfidence and improving the accuracy of the generated outputs. 
Specifically, for a LLM consisting of a total of $L$ layers, we denote the  probability for $y_t \in \mathcal{V}$ in the $l$-th layer as:
\begin{equation}
    p_\theta^l(y_t|x,y_{<t}) = \operatorname{softmax}(\bm{W}_{\textsc{lm}}h^l_{t-1}),
\end{equation}
where $h^l_{t-1}$ denotes the hidden state for layer $l$ out of $L$, and $\bm{W}_{\textsc{lm}}$ denotes the linear classification head in the LLM. At each decoding step,  we dynamically select the layer with the maximum uncertainty for contrast:
\begin{equation}
\begin{aligned}
    l^* & = \arg \max_{l \in \mathcal{L}} H^l_t,\\ 
        H^l_t & = -\sum_{y_t \in \mathcal{V}} p_\theta^l(y_t|x,y_{<t}) \log p_\theta^l(y_t|x,y_{<t}),
\end{aligned}
\label{fml: maxl}
\end{equation}
where $\mathcal{L}$ is the set of candidate layers, which are set as the last few layers of LLMs practically to ensure that each of them contains certain plausible information.
Combining Eq.~\eqref{fml:entropy} and Eq.~\eqref{fml: maxl}, the adjusted next-token distribution is formulated as:
\begin{align}
\label{fml:clehe}
& \! y_t \sim \operatorname{softmax} \Big[ (1\!+\!\beta)\!\!\sum_{d_j \in D} \!w_{j,t}^{\textsc{h}} \log p_\theta (y_t|d_j, \!x, \! y_{<t}) \nonumber \\
& \quad \quad \quad - \beta \log p_\theta^{l^*} (y_t|x,y_{<t}) \Big]. 
\end{align}
Here, $\beta$ represents the amplification intensity of external knowledge. When $\beta = 0$, Eq.~\eqref{fml:clehe} degenerates to the proposed \modela~in \cref{subsec: leens}. 
The ultimate distribution in Eq.~\eqref{fml:clehe} can be interpreted as a two-stage ensemble process. 
Firstly, it ensembles the retrieved documents with uncertainty to generate a low-entropy distribution that more effectively captures the external knowledge within these documents. 
Secondly, it performs a contrastive ensemble by differentiating the logits of this low-entropy distribution from the high-entropy distribution of parametric knowledge selected across different layers, thereby prioritizing factual information from external sources.
In this regard, we term the method in Eq.~\eqref{fml:clehe} as \textbf{\modelb} (\textbf{C}ontrasting \textbf{L}ow-\textbf{e}ntropy distribution with \textbf{H}igh-\textbf{e}ntropy distribution). Figure~\ref{fig:method-illustration} illustrates the overall pipeline of the proposed \modelb.

\section{Related Works}

\paragraph{Retrieval-Augmented Language Models.}
Enhancing large language models (LLMs) with information retrieved from external knowledge bases has proven effective for various knowledge-intensive tasks. 
Initially, mainstream research in retrieval-augmented language models (RALM) focused on leveraging retrieved knowledge during the pre-training phase of LLMs \citep{guu2020retrieval,izacard2023atlas,borgeaud2022improving}.
To mitigate the computational costs, some studies have concentrated on lightweight fine-tuning methods to integrate retrieval capabilities into LLMs \citep{lewis2020retrieval,lin2023ra,zhang2024raft}.
Notably, models like FiD~\citep{izacard2020leveraging} and CEPE~\citep{yen2024long} perform parallel encoding of multiple retrieved documents using a fine-tuned encoder, enabling decoder-only LLMs to more effectively capture and utilize external knowledge.
Another approach leverages the in-context learning abilities of LLMs to incorporate external knowledge in a training-free manner \citep{ram2023context,shi2023replug}. 
The work most closely related to ours is \replug \citep{shi2023replug}, which utilizes the RAG-token model \citep{lewis2020retrieval} to perform parallel retrieval augmentation based on retrieval scores. However, we empirically demonstrate that focusing on the inherent uncertainty within the LLM’s output distribution, rather than relying solely on pre-existing retrieval scores, can significantly improve the factual accuracy of content generated from retrieved documents.

\paragraph{Contrastive Decoding.}
The idea of contrastive decoding (CD) has been previously applied in controllable text generation to produce non-toxic by DExperts \citep{LiuSLSBSC20}. Later, \citet{li2022contrastive} formalized CD as a method to enhance open-ended text generation without any additional training by maximizing the difference in log probabilities between an expert LLM and an amateur LLM. This approach has demonstrated strong performance in various domains, including reasoning~\citep{o2023contrastive} and neural machine translation~\citep{waldendorf2024contrastive}. 
CD can also be interpreted as maximizing pointwise mutual information (PMI), which has proven effective in other scenarios. For instance, \citet{LiGBGD16} uses a training objective that maximizes PMI to generate more diverse conversational responses, while CAD~\citep{shi2023trusting} employs a PMI-adjusting distribution to resolve the knowledge conflict. \citet{chuang2023dola} proposes a decoding strategy that contrasts different layers of the same LLM to more effectively highlight factual knowledge. Similar principles are also applied in visual LLMs, where \citet{leng2023mitigating} mitigates object hallucination by contrasting distributions derived from original and distorted visual inputs. Alternatively, our proposed \modelb~leverages layer-wise entropy-based contrastive decoding to prioritize external knowledge over the parametric knowledge inherent in the LLM itself.

\vspace{-1mm}
\section{Experiments}
\vspace{-1mm}
\paragraph{Baselines.}
We mainly compare three training-free baselines, two of which are ensemble-based methods. \naive: this method concatenates all retrieved documents directly along with the question to form the prompt for LLMs; \replug~\citep{shi2023replug}: it utilizes a normalized retriever weight to ensemble during the decoding process; \avgens~: it follows the formulation in Eq.\eqref{fml: poe} and assigns the same weight to each document during each generation step. In \cref{subsubsec:ablation_study}, we also incorporate two contrastive decoding methods (\ie, CAD~\citep{shi2023trusting} and DoLa~\citep{chuang2023dola}) into our setting and compare with them to validate the effectiveness of our method. Experiments are conducted on 4 LLMs: \textsc{Llama}-2-7B, \textsc{Llama}-2-13B, Mistral-7B-v0.1, and \textsc{Llama}3-8B. Among them, \textsc{Llama}-3-8B supports an 8K context length, while the other three LLMs support a 4K context length.  Few evaluation samples exceed 4k in context length, and they are removed to ensure the validity of the assessment.

\paragraph{Implementation Details.}
Our method introduces two hyperparameters: $\tau$ to control the relative importance of different documents during decoding; and $\beta$ balancing contextual and parametric knowledge. We extract a subset from the WebQ training set for validation to determine the hyperparameter value for each LLM. Ultimately,  $\tau$ is set as 0.25 for \textsc{Llama}-3-8B and as 0.1 for the other three models. For $\beta$, 5.0 is chosen for $\llama$-2-7B, while the other models are assigned a value of 0.25. During the inference, greedy decoding is utilized for reproducibility.  When looking for the layer with the highest entropy, we focus our search exclusively on the candidate layers. In our preliminary experiments, this approach improves computational efficiency and slightly enhances model performance. For \llama-2-7B, Mistral-7B-v0.1, and \llama-3-8B with 32 hidden layers, the candidate layers are set to $\{17, \dots, 32\}$, and only even-numbered layers will be searched. For \llama-2-13B with 40 hidden layers, the candidate layers are set to $\{31, \dots, 40\}$. All experiments are conducted on a single A100 80GB GPU.

\begin{figure*}[!t]
    \centering
    \includegraphics[width=\linewidth]{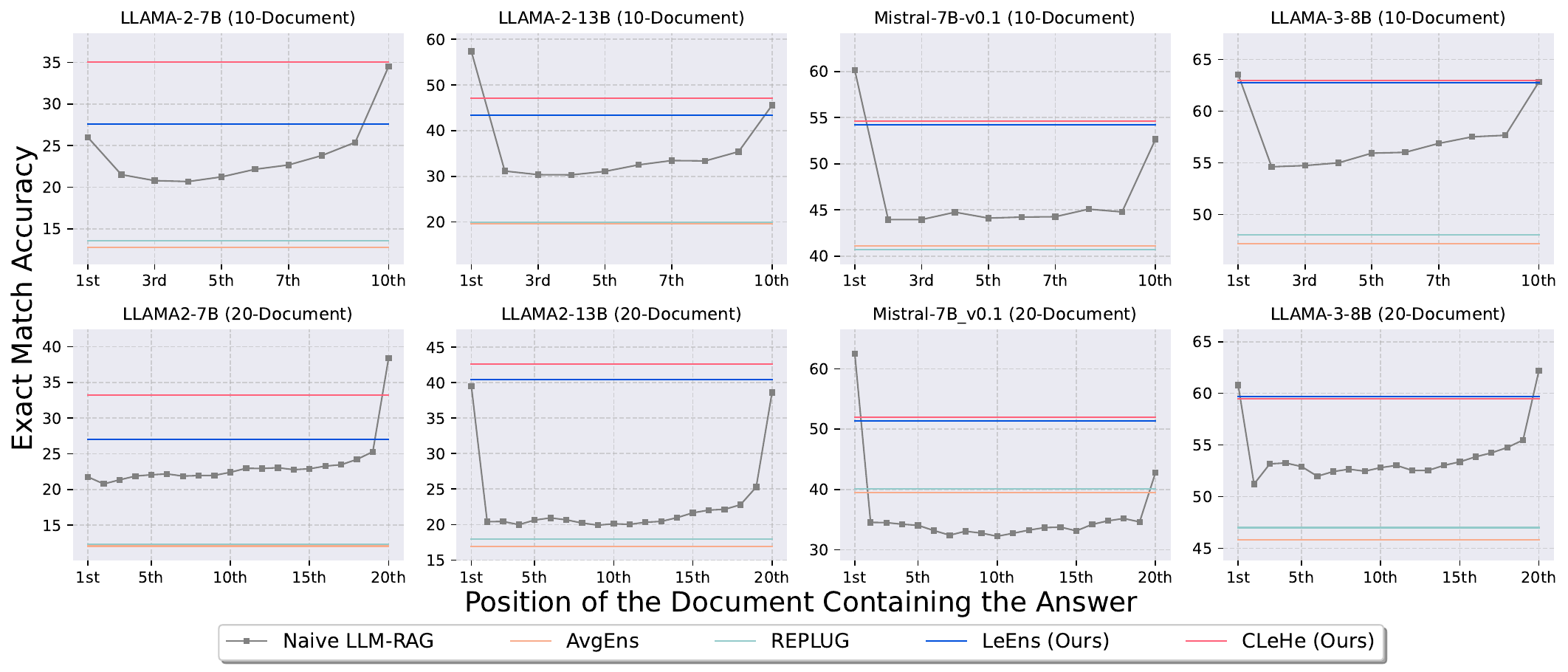}
    \vspace{-4mm}
    \caption{Impact of positioning the oracle document on multi-document question answering performance. A 10-document context typically uses less than 2K tokens; a 20-document context usually uses less than 4K tokens.}
    \label{fig:lost}
    \vspace{-1mm}
\end{figure*}

\begin{figure*}[!t]
  \centering
  \begin{minipage}{0.23\linewidth}
    \centering
    \includegraphics[width=\linewidth]{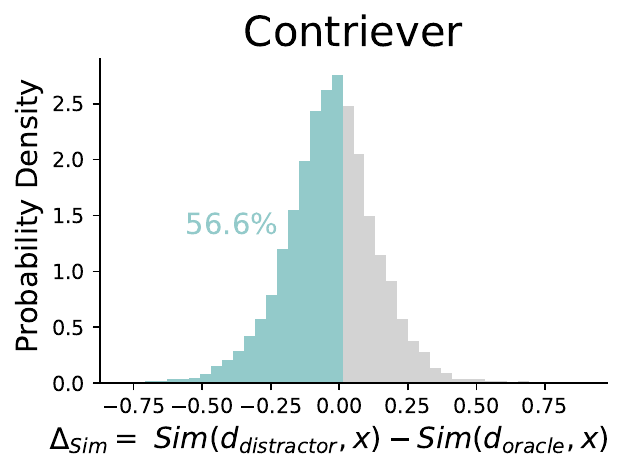}
    \subcaption{}
    \label{fig:score_retrieve}
  \end{minipage}
  \begin{minipage}{0.74\linewidth}
    \centering
    \includegraphics[width=.92\linewidth]{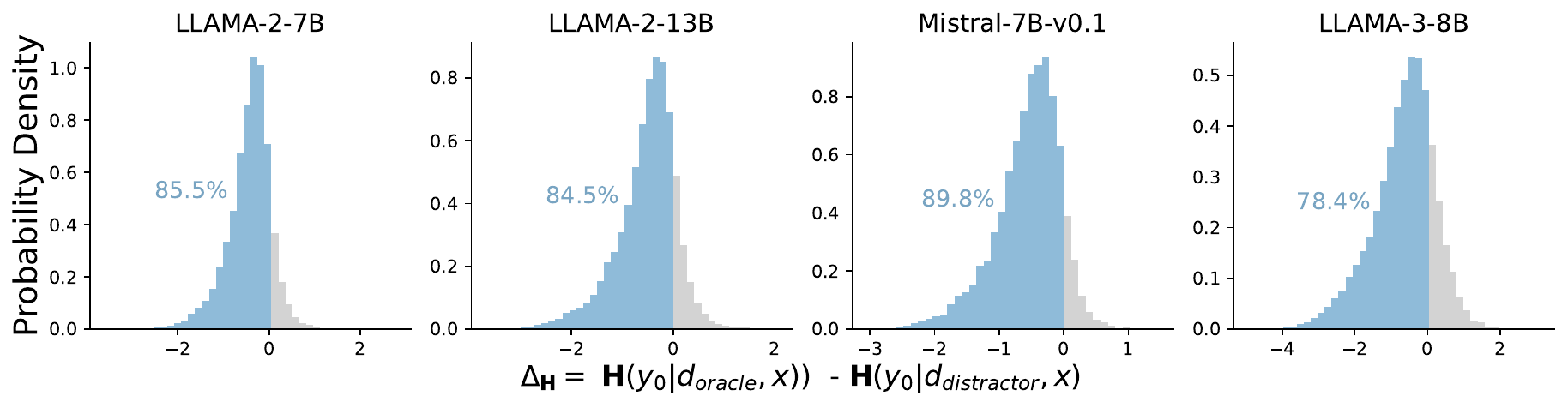}
    \subcaption{}
    \label{fig:score_entropy}
  \end{minipage}
    \vspace{-2mm}
  \caption{(a) The distribution of the similarity difference between the query and the oracle document versus the query and distractor documents. (b) The distribution of the difference in entropy of the first token generated by LLMs when given the oracle document versus distractor documents.}
  \label{fig:score_diff}
  \vspace{-4mm}
\end{figure*}

\vspace{-1.5mm}
\subsection{Analyzing the Distraction Phenomenon in Retrieved Context}
\label{subsec:middle}
We are particularly interested in a challenging QA scenario proposed by ~\citep{liu2024lost}, in which the oracle document is surrounded by numerous semantically similar distractor documents. This configuration challenges the efficacy of the \naive~LLM-RAG, preventing it from accurately deriving answers based on contextual information, thereby leading to the ``Lost in the middle'' phenomenon. Following~\citep{liu2024lost}, given a query from NaturalQuestions-Open~\citep{kwiatkowski2019natural}, we select a Wikipedia paragraph containing the answer from the NaturalQuestions annotations as the oracle document. Then, Contriever~\citep{gautier2022unsupervised} is employed to extract K-1 additional paragraphs from the Wikipedia corpus that are highly relevant to the query yet do not include the ground truth answer, functioning as distractor documents. The query, the oracle document, and K-1 distractor documents are subsequently processed by the LLM to generate an answer.

From Figure~\ref{fig:lost}, it is evident that the performance of the \naive~method which concatenates all documents to compose the context, is highly sensitive to the placement of the oracle document within the retrieved documents. Among the four evaluated LLMs, the performance of the \naive~method significantly deteriorates when the oracle document is neither at the very beginning nor at the end. In contrast, since the proposed \modela~ processes each document in parallel during decoding, its performance is naturally independent of the position of the oracle document. In almost all positions, \modela~substantially surpasses the performance of the \naive~method. Notably, in this challenging scenario, \replug~which ensemble documents’ distributions based on retriever weights perform exceedingly poorly, achieving results merely on par with \avgens~. Based on this observation, we further conduct the weight analysis in Figure~\ref{fig:score_diff}. As depicted in Figure~\ref{fig:score_retrieve}, in only approximately 57\% of instances, Contriever identifies the oracle document as more similar to the query than the distractor documents. Figure ~\ref{fig:score_entropy} shows the distribution of entropy differences for the first token when conditioned on oracle documents versus distractor documents. It indicates that the entropy is generally lower when the response to a query is based on the oracle document rather than the distractor documents. Moreover, the proposed \modelb~derived from \modela~further enhances the accuracy of responses in LLMs, particularly evident in the \llama2-7B and \llama-2-13B. This indicates that contrasting low-entropy distributions of contextual knowledge with high-entropy distributions of parameterized knowledge can further strengthen LLMs' understanding of context.
\begin{table*}[!t]
    \renewcommand{\arraystretch}{1.08}
    \centering
    \small
    \begin{tabularx}{\textwidth}{p{1.6cm} XXX XXX XXX XXX c}
        \toprule
        \textbf{Dataset}    & \multicolumn{3}{c}{\textbf{NQ}}   & \multicolumn{3}{c}{\textbf{TQA}}  & \multicolumn{3}{c}{\textbf{WebQ}} & \multicolumn{3}{c}{\textbf{PopQA}}  &\multirow{2}{*}{\textbf{\normalsize Gain}}  \\
        \cmidrule(lr){2-4} \cmidrule(lr){5-7} \cmidrule(lr){8-10} \cmidrule(lr){11-13} 
        \textbf{\# of Docs} & 5 & 10    &20 & 5 & 10    &20 & 5 & 10    & 20 & 5 & 10    & 20   &~ \\
        \midrule
        \multicolumn{14}{l}{\llama-2-7B} \\
        \hdashline
        \naive  &19.56  &26.76  &26.37 &52.53  &59.05  &66.00  &16.04  &18.06  &18.35  &23.64  &25.50  &26.10    &0.00 \\
        \avgens  &14.93 &13.51  &12.78  &51.04 &49.84  &48.87   &12.50 &12.06  &11.35   &14.03 &12.49  &11.74   &-9.40 \\
        \replug  &23.85  &23.87  &23.92  &55.56 &55.37  &55.31  &16.63  &16.39  &16.77  &23.85  &23.80  &23.91    &-1.56 \\
        \rowcolor{gray!25}
        \modela  &25.48  &25.79  &25.90  &61.87  &62.72  &63.16  &20.28  &21.65  &21.66  &27.84  &26.57 &25.07    &+2.50 \\
        \rowcolor{gray!25}
        \modelb &\textbf{37.62} &\textbf{36.29}  &\textbf{35.48}  &\textbf{69.56}  &\textbf{69.92}  &\textbf{69.72}  &\textbf{36.12}  &\textbf{36.22}  &\textbf{35.77}  &\textbf{32.12}  &\textbf{31.11}  &\textbf{28.89}   &+\textbf{11.74} \\
        \midrule
        \multicolumn{14}{l}{\llama-2-13B} \\
        \hdashline

        \naive  &\textbf{37.98}   &\textbf{39.67}  &29.07  &66.82  &68.15  &69.66  &29.83  &37.02  &27.58  &\textbf{34.77}  &\textbf{35.14}  &31.62    &0.00 \\
        \avgens  &23.52 &20.19  &18.71  &66.92 &65.48  &63.86   &25.89 &24.16  &22.64   &24.28 &22.58  &22.02   &-8.92 \\
        \replug  &34.12 &33.82  &34.01  &67.83  &67.77  &67.60   &31.25  &31.30   &30.88 &30.79  &30.95  &31.00   &-1.33 \\
        \rowcolor{gray!25}
        \modela  &36.54   &34.65  &33.63  &71.87  &72.07  &72.16  &36.07  &35.63  &35.42  &34.63  &33.31  &31.72   &+1.70 \\
        \rowcolor{gray!25}
        \modelb &37.31 &36.01  &\textbf{34.95}  &\textbf{72.24}  &\textbf{72.82}  &\textbf{72.55}  &\textbf{38.19}  &\textbf{37.45}  &\textbf{36.66}  &34.22  &33.16  &\textbf{31.79}    &+\textbf{3.18} \\
        \midrule
        \multicolumn{14}{l}{Mistral-7B-v0.1} \\
        \hdashline
        \naive  &46.20   &44.43  &42.20  &73.53  &70.31  &73.89  &47.69  &45.03  &40.61  &40.23  &37.72  &38.94    &0.00 \\
        \avgens  &40.91 &39.36  &38.22  &76.28  &75.73  &74.91  &47.98  &48.13  &47.83  &37.11  &34.61  &33.44   &-0.52 \\
        \replug  &44.35 &44.44  &44.58  &74.62  &74.80  &74.60  &47.59  &47.49  &47.21  &37.76  &37.75  &37.78   &+1.02 \\
        \rowcolor{gray!25}
        \modela  &\textbf{46.40} &\textbf{46.45}  &\textbf{44.65}  &\textbf{78.26}  &\textbf{78.97}  &\textbf{79.25}  &\textbf{49.21}  &49.70  &50.32  &\textbf{42.12}  &43.31  &43.87   &+\textbf{4.32} \\
        \rowcolor{gray!25}
        \modelb &46.32 &46.07  &44.64  &78.19  &78.88  &79.14  &49.06  &\textbf{49.76}  &\textbf{50.37}  &42.11  &\textbf{43.34}  &\textbf{43.89}    &+4.25 \\
        \midrule
        \multicolumn{14}{l}{\llama-3-8B} \\
        \hdashline
        \naive  &\textbf{52.35}   &\textbf{51.69}  &\textbf{52.33}  &79.92  &81.11  &82.10  &50.49  &50.15  &50.12  &40.92  &42.24  &42.95   &0.00 \\
        \avgens  &47.12 &45.70  &44.51  &81.31  &80.73  &79.78  &51.82  &51.24  &51.01  &38.95  &36.49  &35.07   &-2.72 \\
        \replug  &50.39 &50.33  &50.50  &79.07  &79.16  &78.76  &50.20  &50.79  &50.27  &39.08  &39.24  &39.04   &-1.63 \\ 
        \rowcolor{gray!25}
        \modela  &51.74   &50.53  &49.47  &\textbf{81.80}  &82.68  &83.02  &\textbf{52.17}  &50.98  &51.80  &43.63  &\textbf{44.92}  &\textbf{45.62}   &+1.00 \\
        \rowcolor{gray!25}
        \modelb &52.02    &50.78  &49.67  &81.78  &\textbf{82.84}  &\textbf{83.14}  &51.67  &\textbf{51.62}  &\textbf{52.39}  &\textbf{43.86}  &44.84  &45.57   &+\textbf{1.17} \\
        \bottomrule
    \end{tabularx}
    \caption{Performance (\%) comparison of different ensemble-based methods on benchmark datasets. "Gain" refers to the average absolute improvement (\%) across all datasets and different numbers of retrieved documents when compared to the naive baseline. }
    \label{tab:main_results.}
\end{table*}

\vspace{-2mm}
\subsection{Open-Domain Question Answering}
\vspace{-1mm}
\paragraph{Datasets and Metrics.}
We evaluate our proposed method using four open-domain QA datasets. Natural Questions~\citep{kwiatkowski2019natural}, TriviaQA~\citep{JoshiCWZ17}, WebQ~\citep{berant2013semantic} and PopQA~\citep{MallenAZDKH23}. Natural Questions includes real anonymized queries from Google's search engine. We utilize a filtered test set~\citep{LeeCT19} of 3,610 samples with answers limited to no more than five tokens. TriviaQA comprises trivia question-answer pairs that were scraped from the web. We evaluate its development set containing 7,993 samples. WebQ consists of questions generated through the Google Suggest API, with answers that are entities in Freebase. We use its test set of 2,032 samples for evaluation. PopQA is a novel entity-centric open-domain QA dataset that spans a wide range of entity popularity, emphasizing long-tail knowledge. We utilize its test set which includes 14,267 samples for evaluation. For each dataset, we retain only the questions and their corresponding answers. DPR~\citep{KarpukhinOMLWEC20} is employed to retrieve the top-k passages from the Wikipedia corpus (Dec. 20, 2018) via as evidence documents for each question. Specifically, we report the performance of different decoding methods when retrieving the top-5, top-10, and top-20 documents.
Following~\citep{liu2024lost}, exact match accuracy is utilized for performance evaluation.

\subsubsection{Overall Performance}
Table~\ref{tab:main_results.} presents the overall performance comparison between our proposed method and existing baselines on public benchmark datasets. The results show that when compared to the \naive~method, our entropy-ensemble-based \modela~demonstrates significant average performance improvements across various LLMs, indicating its superior ability to extract useful information from the context. Moreover,  \modela~outperforms \replug~and \avgens~in almost all settings, indicating that using the uncertainty of LLM output distributions for document scoring more effectively facilitates generating answers than static retriever similarity and unweighted averaging. Comparing \modela~with \modelb, we observe that further contrasting the ensemble-based low-entropy contextual distribution with the high-entropy distribution of the parametric knowledge leads to performance improvements, particularly noticeable in \llama2-7B and \llama2-13B. These observations substantiate that the proposed entropy-based decoding mechanism markedly augments the extraction and utilization of contextual information.  Further, on Mistral-7b-v0.1 and \llama-3-8B, CLeHe performs similarly to LeEns, indicating no significant enhancement from the contrastive ensemble. We speculate that these two models are less distracted by parametric knowledge when generating answers.

\subsubsection{Ablation Study}
\label{subsubsec:ablation_study}
\begin{table*}[!t]
\renewcommand{\arraystretch}{1.10}
\centering
\small
\begin{tabularx}{\textwidth}{p{2.9cm} >{\centering\arraybackslash}X >{\centering\arraybackslash}X >{\centering\arraybackslash}X >{\centering\arraybackslash}X >{\centering\arraybackslash}X >{\centering\arraybackslash}X >{\centering\arraybackslash}X >{\centering\arraybackslash}X >{\centering\arraybackslash}X >{\centering\arraybackslash}X >{\centering\arraybackslash}X >{\centering\arraybackslash}X}
\toprule
& \multicolumn{3}{c}{\llama-2-7B}    & \multicolumn{3}{c}{\llama-2-13B}    & \multicolumn{3}{c}{Mistral-7B-v0.1}    & \multicolumn{3}{c}{\llama-3-8B} \\ \cmidrule(lr){2-4} \cmidrule(lr){5-7} \cmidrule(lr){8-10} \cmidrule(lr){11-13}
& NQ    & TQA    & WebQ    & NQ    & TQA    & WebQ    & NQ    & TQA    & WebQ    & NQ    & TQA    & WebQ    \\
\midrule
\naive                   & 19.56 & 52.53 & 16.04 & 37.98 & 66.82 & 29.83 & 46.20 & 73.53 & \textbf{47.69} & \textbf{52.35} & 79.92 & \textbf{50.49} \\
\ w/ JSD (DoLa)                & 38.72 & 68.15 & 35.38 & 27.34 & 45.15 & 21.90 & 46.20 & 74.42 & 46.80 & 52.30 & 81.36 & 49.54 \\
\ w/ Last\_Layer (CAD)        & 38.92 & 65.92 & 30.77 & \textbf{41.52} & \textbf{68.96} & \textbf{33.76} & 44.32 & 70.27 & 41.49 & 51.80 & 79.01 & 46.26 \\
\textbf{\ w/ Entropy}                   & \textbf{41.36} & \textbf{70.32} & \textbf{37.30} & 39.91 & 67.38 & 32.73 & \textbf{46.30} & \textbf{74.59} & 47.05 & 52.28 & \textbf{81.60} & 49.61 \\
\midrule
\textbf{LeEns}          & 25.48 & 61.87 & 20.28 & 36.54 & 71.87 & 36.07 & \textbf{46.40} & 78.26 & 49.21 & 51.74 & \textbf{81.80} & \textbf{52.17} \\
\ w/ JSD (DoLa)                     & 35.84 & 67.11 & 33.75 & 17.22 & 35.17 & 11.02 & 46.29 & 78.19 & 49.16 & 52.04 & 81.73 & 51.82 \\
\ w/ Last\_Layer           & 30.74 & 62.85 & 23.08 & \textbf{38.19} & 71.64 & 38.09 & 45.57 & 75.54 & 48.12 & \textbf{52.60} & 81.36 & 52.01 \\
\textbf{\ w/ Entropy (CLeHe)} & \textbf{37.62} & \textbf{69.56} & \textbf{36.12} & 37.31 & \textbf{72.24} & \textbf{38.19} & 46.32 & \textbf{78.26} & \textbf{49.76} & 52.02 & 81.78 & 51.62 \\ \bottomrule
\end{tabularx}
\vspace{-1.5mm}
\caption{Performance on combining different external and parametric knowledge modeling methods. Experiments are conducted under the top-5 document setting.}
\vspace{-4.5mm}
\label{tab:ablation_study}
\end{table*}

Within the contrastive decoding framework, we investigate the compositional effects on performances by combining different modeling techniques for external and parametric knowledge. To extract knowledge from retrieved external documents, we explore two modeling approaches.: \naive~RAG and our entropy-based document ensemble modeling (LeEns). Additionally, we explore three layer-based strategies to derive parametric knowledge: (i) Last-Layer strategy. It defines parametric knowledge using the distribution from the last layer of LLMs when without retrieved context.  CAD~\citep{shi2023trusting} utilizes this strategy, \ie, contrasting the distribution derived from \naive~RAG against the last-layer context-free distribution. (ii) JSD-based strategy. It first calculates the Jensen-Shannon Divergence (JSD) between the RAG-derived distribution and the distribution of each layer when without retrieved documents, then selects the layer with the highest JSD for contrast. (iii) Our proposed entropy-based strategy. It directly selects the layer with the highest entropy as the proxy of intrinsic knowledge.  As shown in Table~\ref{tab:ablation_study}, compared to the other two layer selection strategies, the proposed entropy-based strategy consistently and significantly enhances model performance in both external knowledge modeling ways of ~\naive~RAG and our LeEns. 
This improvement is particularly notable in ~\llama-2-7B. The last-layer strategy markedly impairs the performance of some LLMs such as Mistral-7B-v0.1. Moreover, in our setting, the JSD-based strategy for contrast is found unstable and results in severe performance degradation in \llama-2-13B. Appendix~\ref{app:relations} displays the average entropy of tokens from various layers, indicating that higher entropy layers often yield better performance. Additionally, Appendix~\ref{appendix-beta} details the hyper-parameter needed to replicate Table~\ref{tab:ablation_study}.

\vspace{-1.5mm}
\subsubsection{Hyper-Parameter and Latency Analysis}
\begin{figure}
    \centering
    \includegraphics[width=\linewidth]{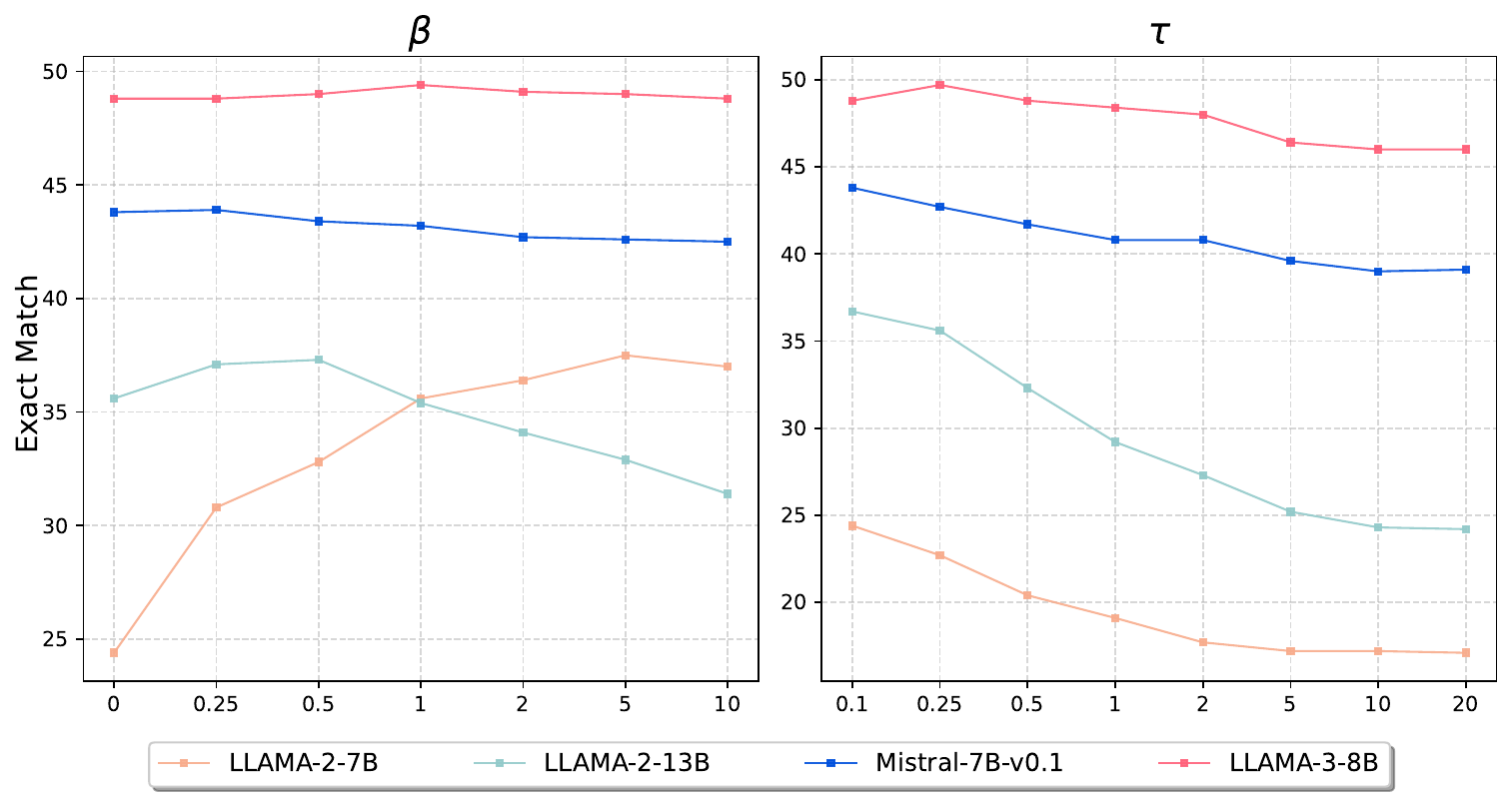}
    \caption{Hyper-parameter analysis using 1K evaluation samples of NQ under the top-5 document setting.}
    \label{fig:hyperprameter}
\end{figure}

We study the influence of the introduced hyperparameters: $\tau$ and $\beta$. As shown in Figure~\ref{fig:hyperprameter}, a small value of $\tau$ (\eg, 0.1 or 0.25) typically results in better performance; as $\tau$ increases, the performance gradually declines. Ideally, when $\tau \rightarrow \infty$,  the performance of the proposed \modela~will match that of \avgens. Regarding $\beta$, it’s observed that for \llama-2-7B, a high $\beta$ (e.g., 5) enables it to effectively contrast the differences between external and parametric knowledge for improved performance. For other evaluated LLMs, we suggest setting it to a small value, saying [0.25, 0.5].

\begin{table}[!t]
    \renewcommand{\arraystretch}{1.18}
    \setlength{\tabcolsep}{4pt}
    \centering
    \small
    \begin{tabularx}{\linewidth}{lccc}
    \toprule
    \textbf{\# of Docs} &\textbf{5} &\textbf{10} &\textbf{20} \\
    \midrule
    \naive  &27.51 ($\times$1.00)  &29.49 ($\times$1.00)    &32.56 ($\times$1.00) \\
    LeEns   &30.43 ($\times$1.11)   &32.90 ($\times$1.12)   &37.76 ($\times$1.16) \\
    CLeHe   &31.88 ($\times$1.16)   &34.10 ($\times$1.16)   &38.49 ($\times$1.18) \\
    \bottomrule
    \end{tabularx}
    \vspace{-1.5mm}
    \caption{Decoding latency (ms/token) of \llama-2-7B based on the number of retrieved documents as context.}
    \label{tab:decoding_latency}
    \vspace{-4mm}
\end{table}
As for decoding latency, Table~\ref{tab:decoding_latency} shows that compared to the \naive~method, our LeEns and CLeHE increase the decoding time by factors of less than 1.18, indicating that they can be applied at a reasonable cost. Appendix~\ref{appendix:decoding_latency_13b} shows the latency of \llama-2-13B that exhibits a similar cost trend.

\vspace{-2mm}
\section{Conclusions}
\vspace{-1mm}
In this paper, we proposed a novel decoding method that is guided by entropy considerations to mitigate the distractibility issue from both external retrieved documents and parametric knowledge. First, we conducted parallel retrieval augmentation with entropy-based ensemble weight to obtain the low-entropy distribution of context. Furthermore, we contrasted this distribution against the highest-entropy distribution among layers when without context to amplify the external knowledge preserved in context. Extensive experiments showed the proposed method's effectiveness in retrieval-augmented open-domain question answering.

\section{Limitations}
One limitation of our study is that we only validated the effectiveness of our method on question answering datasets, without testing it on other knowledge-intensive tasks such as fact verification. Extending the method proposed in this paper to other retrieval-augmented scenarios will be a future research direction. Additionally, due to computational power constraints, we only tested the effectiveness of the proposed method on models with fewer than 13B parameters. However, whether the method proposed in this paper is applicable to LLMs with more parameters (\eg, 70B or more) remains to be explored in future research.

\section*{Acknowledgements}
The work described in this paper was partially supported by Laboratory for AI-Powered Financial Technologies, InnoHK initiative and The Government of the HKSAR. The work described in this paper was also partially supported by the Research Grants Council of the Hong Kong Special Administrative Region, China (CUHK 2410072, RGC R1015-23).

\bibliographystyle{acl_natbib}
\bibliography{custom}

\appendix

\section{Appendix}
\begin{figure*}[!t]
    \centering
    \includegraphics[width=\linewidth]{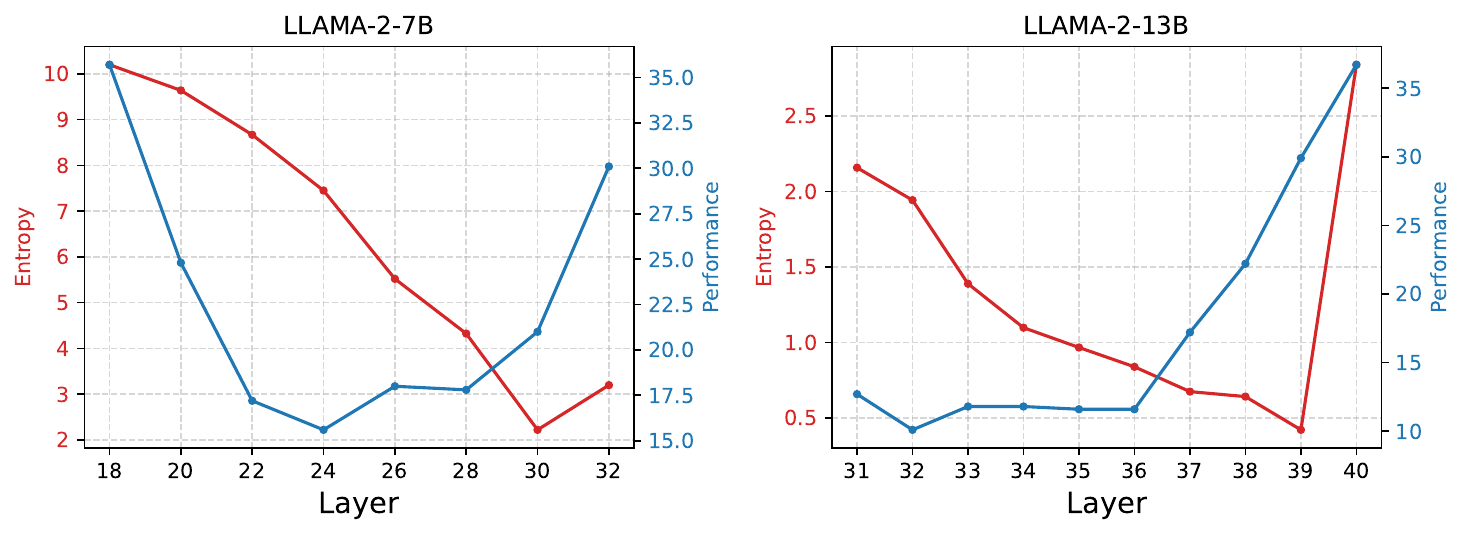}
    \caption{Average entropy of generated tokens and performance for each layer.}
    \label{fig:entropy-performance}
\end{figure*}

\subsection{Additional Implementation Details}
\label{appendix-beta}
As shown in Table~\ref{tab:beta}, for both the dynamic JSD-based selection strategy and the proposed entropy-based strategy, we set the $\beta$ parameter to 5.0 for llama-2-7B and $\beta$ to 0.25 for other LLMs. This configuration ensures that contrastive decoding achieves superior performance across different LLMs. For the last-layer strategy, we found that setting a higher beta significantly severely reduces the performance of LLMs. Therefore, for this strategy, we set $\beta$ to 0.25 for all LLMs.
\begin{table}[h]
\centering
\begin{tabular}{@{}ll@{}}
\toprule
Method               & $\beta$     \\ \midrule
\naive~w/ JSD         & 5.0 / 0.25 \\
\naive~w/ Last\_Layer & 0.25     \\
\naive~w/ Entropy     & 5.0/0.25 \\
LeEns w/ JSD         & 5.0/0.25 \\
LeEns w/ Last\_Layer  & 0.25     \\
LeEns w/ Entropy     & 5.0/0.25 \\ \bottomrule
\end{tabular}
\caption{$\beta$ for different layer selection strategies.}
\label{tab:beta}
\end{table}

\subsection{Relations between Entropy and Performance}
\label{app:relations}
Figure~\ref{fig:entropy-performance} shows the average entropy values of tokens generated on the WebQ dataset by \llama2-7B and \llama2-13B when using different layers for contrast. For \llama2-7B, the entropy initially shows a decreasing trend and later an increasing trend among the tested layers, which correlates somewhat with the overall performance trend. For \llama2-13B, it was found that both the highest performance layer and the layer with the highest entropy are the last layer. This might explain why the performance of our entropy maximization-based layer selection is overall consistent with the performance of the last layer.

\subsection{Illustration of the False Positive Case}
\begin{figure*}[!t]
    \centering
    \includegraphics[width=\linewidth]{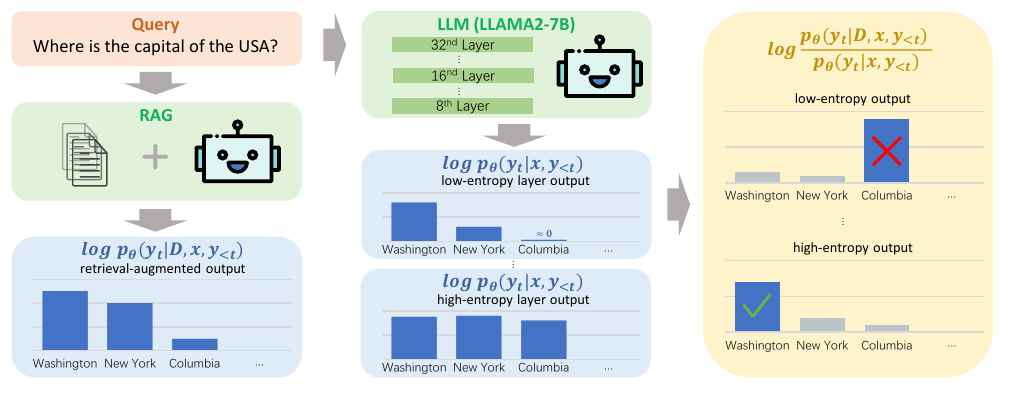}
    \caption{Illustration of the false positive case.}
    \label{fig:app-false-positive}
\end{figure*}

Figure~\ref{fig:app-false-positive} illustrates a false positive scenario. In this example, the retrieval-augmented model assigns high probabilities to the words "Washington", "New York", and "Columbia" as candidate positives. However, in the low-entropy output of a specific layer (typically the last layer) without context, the probability assigned to "Columbia" is notably low. If contrastive decoding is applied, it would mistakenly increase "Columbia's" probability, leading to an incorrect prediction.

\subsection{Prompt Format}

\begin{table}[ht]
    \centering
    \small
    \begin{tabular}{>{\raggedright\arraybackslash\tt}p{0.94\linewidth}<{}}
        \toprule
            Write a high-quality answer for the given question
using only the provided search results. \\
            \\
            \vspace{-1em}
            Document {[1]} (Title: ...) ... \\
            Document {[2]} (Title: ...) ... \\
            Document {[3]} (Title: ...) ... \\
            ... \\
            \\
            \vspace{-1em}
            Question: \{Question\} \\
            Answer: \\
        \midrule
        \midrule
            Write a high-quality answer for the given question
using only the provided search results. \\
            \\
            \vspace{-1em}
            Document (Title: ...) ... \\
            \\
            \vspace{-1em}
            Question: \{Question\} \\
            Answer: \\

        \bottomrule
    \end{tabular}
    \caption{The Prompt format for \naive~RAG (Top); and the prompt format for the proposed \modelb~ (Bottom) to process each document in parallel.
    }
    \vspace{-10pt}
    \label{tab:prompt_format}
\end{table}

In Table~\ref{tab:prompt_format} (Bottom), we showcase the prompt format utilized for the proposed \modelb; its structure is basically the same as that of the \naive~RAG. The primary distinction is that due to the parallel processing, we need to repeat the task prompt and question for each document.

\subsection{Decoding Latency of \llama-2-13B}
\label{appendix:decoding_latency_13b}
\begin{table}[h]
    \renewcommand{\arraystretch}{1.18}
    \setlength{\tabcolsep}{4pt}
    \centering
    \small
    \begin{tabularx}{\linewidth}{lccc}
    \toprule
    \textbf{\# of Docs} &\textbf{5} &\textbf{10} &\textbf{20} \\
    \midrule
    \naive  &35.51 ($\times$1.00)  &37.81 ($\times$1.00)    &49.89 ($\times$1.00) \\
    LeEns   &39.14 ($\times$1.10)   &43.54 ($\times$1.15)   &58.84 ($\times$1.18) \\
    CLeHe   &40.24 ($\times$1.13)   &45.28 ($\times$1.20)   &60.79 ($\times$1.22) \\
    \bottomrule
    \end{tabularx}
    \caption{Decoding latency (ms/token) of \llama-2-13B based on the number of retrieved documents as context.}
    \label{tab:decoding_latency_13b}
\end{table}
As shown in Table~\ref{tab:decoding_latency_13b}, when our methods are applied to the 13B model, the increase in decoding time is capped at factors of 1.22. This increase is considered acceptable given the superior performance our methods deliver.

\end{document}